\pdfoutput=1
\documentclass[oneside]{svproc}

\usepackage{url}
\usepackage{amsmath}
\usepackage{graphicx}
\usepackage{xcolor}
\usepackage{enumitem}
\usepackage[hidelinks]{hyperref}

\begin{document}
\mainmatter

\title{The Language of Deformation:\\ Semantic Failure Analysis as a Prerequisite for Dexterous Manipulation of Deformable Objects}

\titlerunning{The Language of Deformation}

\author{Ignacio Cuiral-Zueco\inst{1} \and David Hardman\inst{2} \and \\ Michele Pierallini\inst{3} \and
Erfan Shahriari\inst{4}}

\authorrunning{Ignacio Cuiral-Zueco\and David Hardman\and Michele Pierallini \and Erfan Shahriari}

\tocauthor{Ignacio Cuiral-Zueco, David Hardman, Michele Pierallini, Erfan Shahriari}

\institute{
Clermont Auvergne INP, Institut Pascal, F-63000 Clermont-Ferrand, France. \email{ignacio.cuiral\_zueco@sigma-clermont.fr}
\and
Department of Engineering,
University of Cambridge
Cambridge CB2 1PZ, UK
\email{dsh46@cam.ac.uk}
\and
Centro di Ricerca ``Enrico Pi\-ag\-gio'', U\-ni\-ver\-si\-t\`{a} di Pisa, 
Largo Lucio Lazzarino 1, 56126 Pisa, Italy. \email{michele.pierallini@gmail.com}
\and
RAI Institute, Zurich 8050, Switzerland,
\email{eshahriari@rai-inst.com}}

\maketitle

\begin{abstract}
Dexterous manipulation of rigid objects has reached a level of maturity where grasp stability, control, and planning strategies are successfully applied in industry and logistics. In contrast, manipulating deformable objects (cables, plastic pieces, cloth) remains far more challenging, as rigid-object methods do not account for their additional complexity. This paper identifies \textit{semantic failure analysis} as a key step toward mature deformable-object manipulation and supports this claim with proof-of-concept experiments. Unlike rigid-object manipulation, where goals are well-defined in terms of criteria such as force balance and slippage, deformable-object manipulation involves a broader variety of failure modes arising from differences in materials, geometry, and behaviour. Typical failures include buckling, plastic deformation, pressure variations, and tearing. These diverse and often hard-to-detect events require distinct corrective strategies, making semantic failure analysis necessary. We define its key components and challenges and present experimental results supporting its potential as a research direction.
\end{abstract}

\section{Introduction}

A driver would not react to a brake failure in the same way as to a skid on ice. The difference lies in the semantic understanding of each event: brake failure (a sinking pedal) demands pumping the brakes, while traction loss (locked wheels) requires the \textit{opposite} action of easing off them. The same principle applies to deformable-object manipulation, where failure modes differ widely in their characteristics and behaviour, each requiring a different — and sometimes non-intuitive — response.

For example, the corrective action needed to maintain a grasp on an object undergoing plastic compression differs sharply from that required when an object is about to buckle. In the first case, maintaining a tight grasp is essential, since even a slight reduction in compression may cause the \textit{permanently} deformed object to slip. In the buckling case, however, the system should reduce compressive forces to prevent a sudden snap that releases the accumulated elastic energy. A capable manipulator must therefore distinguish these failure modes and respond accordingly.

By contrast, in rigid-object manipulation, explicit semantic reasoning has been less critical. Rigid-object behaviour defines a bounded problem space in which tasks, constraints, and corrective actions are described by well-established frameworks, and many strategies generalize across objects. As a result, the system itself rarely needs to classify diverse failure types.

In this paper, we argue that, given the greater variety and complexity of failure modes in deformable-object manipulation, semantic failure analysis is a necessary step toward robust and practical manipulation. After reviewing related work in Sec.~\ref{sec:relatedwork}, we define Semantic Failure Analysis for Deformable Object Manipulation (SFA-DOM) in Sec.~\ref{sec:sfa} and identify the research challenges it entails. Sec.~\ref{sec:experiments} describes proof-of-concept experiments that motivate SFA-DOM and support the identified challenges, and Sec.~\ref{sec:conclusions} summarizes our findings and future directions.

\section{Related Work}\label{sec:relatedwork}
The two most common approaches to generating grasp motions for rigid objects are classical analytical contact models, see, e.g., \cite{pollayil2022sequential,prattichizzo2013motion,gabiccini2012grasp}, and data-driven learning-based policies, see, e.g., \cite{bohg2013data,mahler2016dex,palleschi2023grasp,murali2025shared,voigt2025learning}.
The former relies mainly on analytical rigid-body contact mechanics and requires exact knowledge of the coefficient of friction, as well as precise real-time force/torque measurements \cite{pollayil2022sequential,gabiccini2012grasp}.
The latter combines a demonstration-driven initial guess of the grasping primitive with simulation-based or model-based refinement \cite{bohg2013data}. These approaches can be computationally expensive due to their reliance on visual and tactile data. However, recent works show that these pipelines can achieve robust grasping from one or a few demonstrations \cite{palleschi2023grasp}, generalize to unseen objects \cite{murali2025shared}, and operate in real-world scenarios~\cite{voigt2025learning}.
Grasps are then executed by closing the loop on high-rate tactile and force/torque feedback \cite{dutta2024vitract}. In particular, tactile sensors or sensorized soft hands detect slip, which the policy or controller then compensates for \cite{voigt2025learning,gilday2023predictive}. These methods extend to the manipulation of deformable objects \cite{li2024grasping}. However, the internal deformation, the non-homogeneous material response, and the hard-to-model dynamics of deformable objects pose fundamental challenges \cite{yuanyang2025grasping}. Simulations of deformable objects exhibit a sim-to-real gap \cite{heckers2024let,zafra2025survey}, highlighting the need for data-efficient strategies, which remain an open problem \cite{palleschi2023grasp,zhang2025particle}. Further, with vision alone, the state of deformable objects is only partially observable \cite{navarro2013model}. Thus, tactile control loops must incorporate sensors and strategies that minimize latency and frequency mismatch.

\section{Semantic Failure Analysis for Deformable Object Manipulation (SFA-DOM)} \label{sec:sfa}

\begin{figure}[t]
    \centering
    \includegraphics[width=1\linewidth]{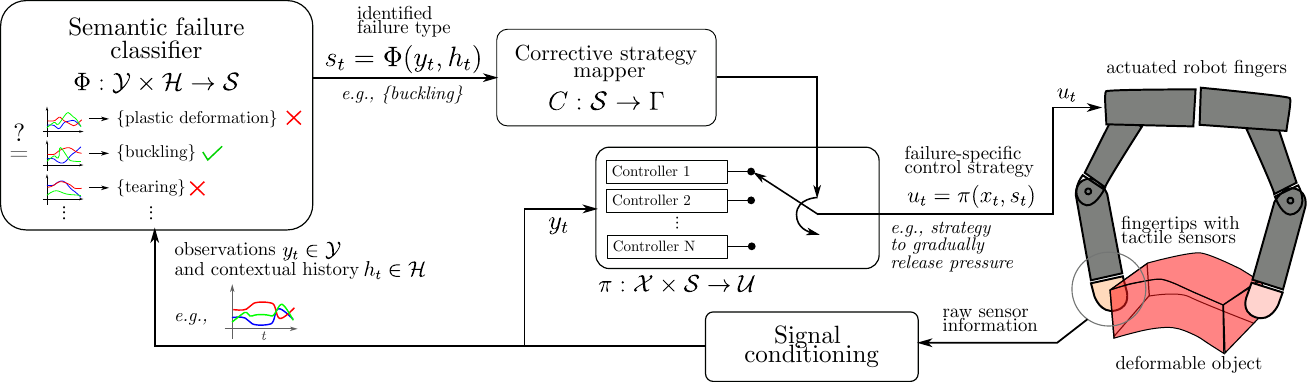}
    \caption{Schematic representation of the SFA-DOM framework. Sensor observations $y_t$ and contextual history $h_t$ are processed by the \textit{semantic failure classifier} $\Phi$ to infer a semantic failure label $s_t$, which the \textit{corrective strategy mapper} $C:\mathcal{S}\rightarrow\Gamma$ then uses to select an appropriate corrective response. Combined with the \textit{control policy} $\pi:\mathcal{X}\times\mathcal{S}\rightarrow\mathcal{U}$, this selection yields failure-aware control actions $u_t$. The depicted tactile sensors, signal conditioning, and controller selection are purely illustrative.}
    \label{fig:mainflowchart}
\end{figure}

This section defines Semantic Failure Analysis for Deformable Object Manipulation (SFA-DOM) at a high level (Fig.~\ref{fig:mainflowchart}) and identifies the main research challenges it poses.

\subsection{General SFA-DOM definition}

\begin{definition}[Semantic Failure Analysis for Deformable Object Manipulation (SFA-DOM)]
\normalfont
\label{def:sfa_dom}
Let a deformable object be modeled as a dynamical system
$O = (\mathcal{X},\mathcal{U},\mathcal{Y},f)$,
where $\mathcal{X}$ is the state space, $\mathcal{U}$ the control input space, $\mathcal{Y}$ the observation space, and
$f:\mathcal{X}\times\mathcal{U}\rightarrow\mathcal{X}$ the state transition function.
At time $t$, the system produces an observation $y_t\in\mathcal{Y}$, and the contextual history $h_t\in\mathcal{H}$ is the finite sequence of recent observations and control inputs $h_t = (y_{{t-k}:{t-1}}, u_{{t-k}:{t-1}})$.

Semantic Failure Analysis (SFA) formalizes the process of interpreting observations to determine the type of failure and to drive corrective planning or control. Concretely, SFA constructs a semantic classifier
\begin{equation}
\Phi:\ \mathcal{Y}\times\mathcal{H}\ \rightarrow\ \mathcal{S},\qquad s_t=\Phi(y_t,h_t),
\label{eq:SFA_problem}
\end{equation}
where $\mathcal{S}$ is a finite (or structured) set of semantic failure classes (e.g., buckling, plastic deformation, tearing), characterized by their causal mechanism, observable signature, and the class of control responses they imply. Given a semantic label $s_t$, SFA supports the selection of an appropriate corrective or preventive strategy. This selection can be represented as a mapping
\begin{equation}
C:\ \mathcal{S}\ \rightarrow\ \Gamma,
\label{eq:C}
\end{equation}
where $\Gamma$ denotes a library of corrective/planning primitives, or as a closed-loop policy that conditions control on the semantic state:
\begin{equation}
\pi:\ \mathcal{X}\times\mathcal{S}\ \rightarrow\ \mathcal{U},\qquad u_t=\pi(x_t,s_t).
\label{eq:controllers}
\end{equation}
Thus, SFA-DOM is the construction and application of $\Phi$ (together with the downstream mapping $C$ or policy $\pi$) so that sensor observations are mapped to meaningful failure labels that inform corrective decisions for robust manipulation of deformable objects.
\end{definition}

\subsection{SFA-DOM: main challenges and implications}

Given Def.~\ref{def:sfa_dom}, the main challenges and implications of realizing SFA-DOM fall into four areas:
\begin{enumerate}
    \item \textbf{Sensing:} Two objects with identical shape and grasp configuration can exhibit very different physical behaviours, resulting in distinct failure modes (e.g., plastic deformation vs.\ buckling) that often become apparent only after the failure has progressed too far. Vision-based systems are generally insufficient because of occlusions and the lack of detailed force feedback. Tactile sensing at the manipulation surface is therefore needed to continuously identify failure events from time-series data. Temporal resolution is critical: many failures involve large changes that occur over short periods or in response to \textit{very} small inputs. SFA-DOM thus relies on precise tactile feedback at sufficiently high sampling rates. Fine contact perception need not rely solely on surface tactile sensors: robots with low reflected inertia, high-bandwidth torque control, and calibrated proprioceptive sensing can also estimate contact dynamics from internal force or current feedback. However, such capabilities remain rare in conventional industrial manipulators, where high gear ratios and joint friction mask subtle interaction forces.
    
    \item \textbf{Actuation:} Dexterous interaction with deformable objects requires force-aware actuation capable of producing and regulating forces at the scale of the interaction — often a few newtons or less. This requires actuators with low reflected inertia, tunable compliance, and high backdrivability for smooth, fine-resolution force generation. Because contact dynamics can change abruptly as geometry or stiffness evolves, high mechanical bandwidth and low latency are also essential. Repeatable actuation at such small force scales is further needed to produce reliable datasets for training and validating SFA-DOM models.
    
    \item \textbf{Modeling:} Accurately modeling deformable objects remains a central open problem: no widely accepted formulation accounts for the non-linear, history-dependent behaviour of soft materials in a way that is both analytically tractable and computationally efficient. When available, such models can support SFA-DOM either analytically, to derive semantic failure descriptors, or in simulation, to generate synthetic training scenarios. However, existing approaches face a fidelity--cost trade-off: high-accuracy models are too slow for large-scale data generation, while efficient models sacrifice fidelity. Consequently, SFA-DOM cannot rely solely on simulation and requires real, high-quality interaction data, reinforcing the need for rich sensing and precise, repeatable actuation.
    
    \item \textbf{Control and Reactivity:} Once semantic failure states are identified, the robot must react by selecting a corrective primitive from $\Gamma$ via the mapping $C$ (Eq.~\ref{eq:C}) or adjusting its control through the policy $\pi$ (Eq.~\ref{eq:controllers}). This requires deriving failure-specific control strategies, whether through model-based formulations (e.g., trajectory constraints, impedance modulation) or data-driven learning conditioned on semantic states. In both cases, the usefulness of SFA-DOM depends on the robot's actuation bandwidth and stability: if force resolution or control rate is insufficient, corrections become coarse and feedforward rather than responsive. Thus, the framework is practically useful only if it accounts for the coupling between semantic reasoning and physical execution.
    
\end{enumerate}

\section{SFA-DOM Proof-of-Concept Experiments: Failure Case Analysis and Signatures with Fine Tactile Sensing}\label{sec:experiments}

\begin{figure}[!h]
    \centering
    \includegraphics[width=1\linewidth]{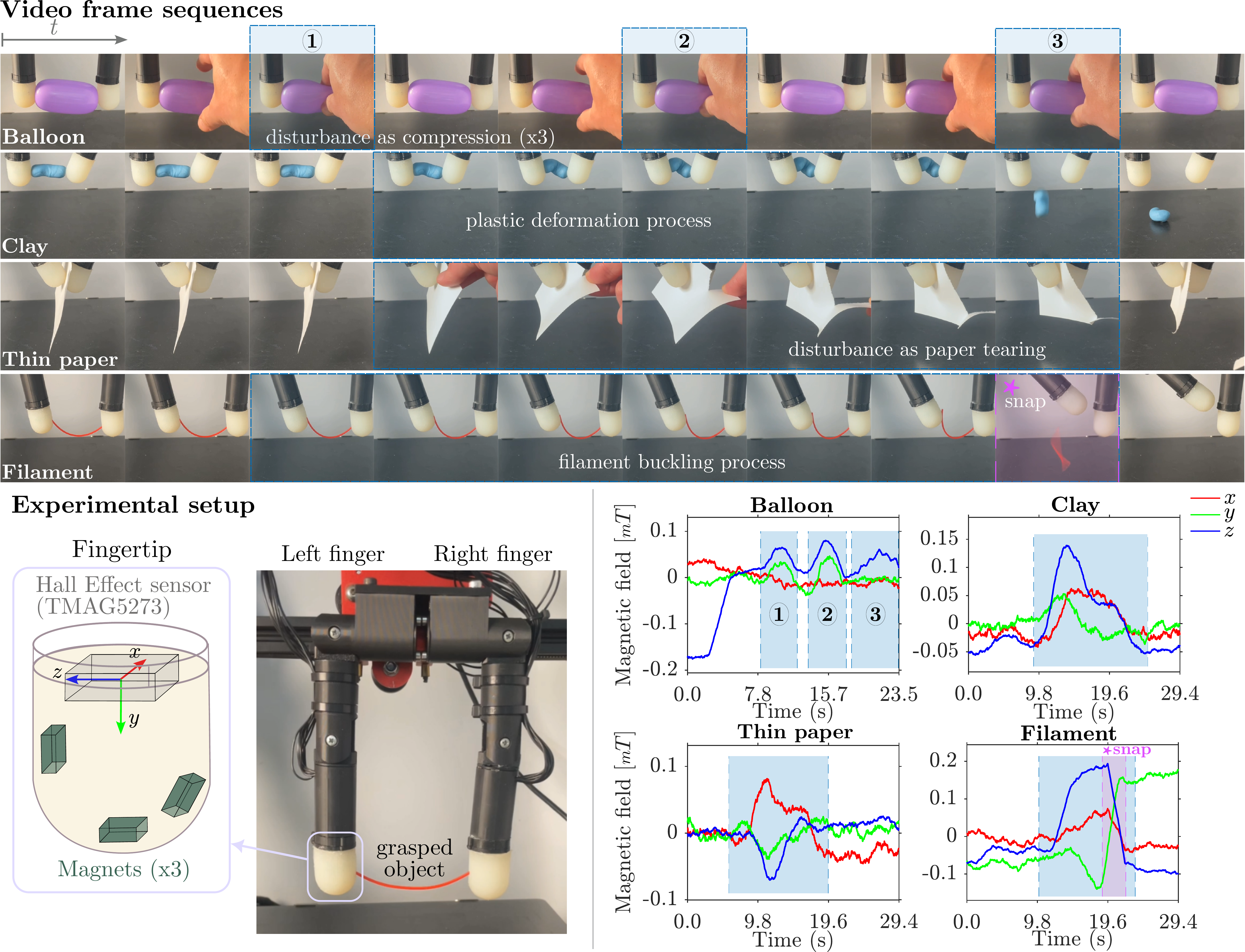}
    \caption{Experimental setup and tactile sensing results for interactions with deformable and fragile objects. 
({Top}) Video frame sequences showing manipulation of a balloon, a clay piece, a thin paper sheet, and a filament, illustrating distinct events: compression, plastic deformation, tearing, and buckling.
({Bottom left}) Sensing configuration.
({Bottom right}) Recorded magnetic-field components $x$, $y$, and $z$ (R, G, B) for each object interaction. Blue shaded regions highlight characteristic tactile signatures corresponding to each event: three compressions (balloon), plastic deformation and grip loss (clay), tearing (paper), and buckling with snapping (filament).}
    \label{fig:experiments}
\end{figure}

This section presents preliminary experiments that empirically motivate SFA-DOM (Def. \ref{def:sfa_dom}) and support the challenges identified in Sec.~\ref{sec:sfa}. Each experiment in Fig.~\ref{fig:experiments} induces a specific failure mode in a deformable object while recording tactile data from a Hall-effect sensor (TMAG5273) embedded in a robot finger, as described in \cite{hardman2025automated}. The sensor measures magnetic-field variations induced by the motion of three internal magnets during fingertip deformation. Distinct interaction events yield clearly distinguishable tactile signatures $y_t$, supporting the feasibility of semantic classifiers $\Phi$ (Eq.~\ref{eq:SFA_problem}).

The observed events caused magnetic-field variations of $\Delta B \approx 0.1~\mathrm{mT}$, which, based on the calibrated sensitivity\footnote{This represents an upper-bound conversion ratio, as the calibration was performed assuming a single point contact between the object and the robot finger, whereas some experiments involve a non-negligible contact area.},
\begin{equation}
\frac{\Delta B}{\Delta F} \approx 0.136~\mathrm{mT/N},
    \label{eq:mTNconversion}
\end{equation}
correspond to contact-force changes on the order of $\Delta F \approx 0.1/0.136~\mathrm{N} \approx 0.74~\mathrm{N}$.  
These tactile signatures, ranging from the \textit{pulses} of balloon compression, to the non-linear decay of clay plastic deformation, to the abrupt high-frequency transient of filament buckling, motivate the four key SFA-DOM challenges introduced in Sec.~\ref{sec:sfa}:

\begin{enumerate}
    \item \textbf{High-Resolution Contact Sensing:}
    High-bandwidth tactile data are critical for capturing subtle, rapid events such as the \textit{fast filament snap} (around 19~s) or the \textit{non-linear force decay} during clay deformation (around 15~s), phenomena often missed by vision or low-resolution sensors. Using the conversion in Eq.~\ref{eq:mTNconversion}, clay plastic deformation ($\Delta B \approx 0.1~\mathrm{mT}$) requires a force-sensing resolution of $\Delta F \approx 0.74~\mathrm{N}$ or finer, while the filament snap ($\Delta B \approx 0.03~\mathrm{mT}$) implies $\Delta F \approx 0.22~\mathrm{N}$. These magnitudes approach the noise floor of commercial wrist-mounted force/torque sensors such as the FT~300-S ($\approx 0.1~\mathrm{N}$ per axis), which are further limited by mechanical compliance and transmission losses between the wrist and the contact point. Proprioceptive sensing (joint-torque or current feedback) faces similar limitations, as modeling inaccuracies and sensor noise can dominate these sub-newton signals. High-resolution, contact-proximal tactile sensing, such as the embedded TMAG5273 configuration in Fig.~\ref{fig:experiments}, is therefore essential.
    
    \item \textbf{Low-Inertia Fine Force Actuation:}
    Managing events such as those in Fig.~\ref{fig:experiments} (preventing filament buckling or inducing repeatable clay deformation for training) requires smooth sub-newton force control ($\leq 0.5~\mathrm{N}$). High-gear-ratio manipulators such as the Franka FR3 or UR5e exhibit effective endpoint masses well above those of a human arm ($\sim$1--4~kg) \cite{okamura2000overview} due to their harmonic-drive transmissions \cite{haddadin2007safety}, which masks fine interaction forces. Series-elastic actuators (SEAs) can reduce apparent stiffness, but their compliance limits mechanical bandwidth. Direct- and quasi-direct-drive architectures offer a better compromise, combining low gear ratios with high torque fidelity for smooth, backdrivable force output. SFA-DOM thus requires backdrivable, low-inertia mechanisms with precise force control.

    \item \textbf{Data-Driven Deformable Object Modeling:}
    The complex, non-linear signal patterns (e.g., the decay during plastic deformation of the clay piece, or the non-linear asymmetric response of the $x$ and $z$ components during paper tearing) illustrate the limitations of first-principles modeling and the need for data-driven models. Physics-based approaches such as the Finite Element Method (FEM) or Cosserat rod theory suffer from a critical sim-to-real gap. This gap arises primarily from the difficulty of calibrating their physical parameters (e.g., Young's modulus, damping, plasticity). Such system identification is a notoriously hard inverse problem, often requiring complex, data-intensive, offline optimization, and remains a major open research challenge \cite{arriola2020modeling}. Consequently, even after calibration, the resulting models exhibit limited force accuracy (e.g., RMSE of $\approx$1.5 N \cite{zhu2020semi}), an order of magnitude too coarse for regulating the sub-newton events in Fig.~\ref{fig:experiments}.

    \item \textbf{Failure-Specific Reactive Control:}
    While the observed tactile signatures might seem counterintuitive, they are visually distinguishable, which suggests that semantic labels $s_t$ can be reliably defined and obtained. Therefore, given the three prerequisites discussed above (fine sensing, fine force control, and precise data-driven models), learning the failure-specific corrective mapping $C$ or control policies $\pi$ (Eq.~\ref{eq:controllers}) within an SFA-DOM framework is plausible.
\end{enumerate}

\section{Conclusions}\label{sec:conclusions}
We propose Semantic Failure Analysis for Deformable Object Manipulation (SFA-DOM) as a research direction for robust robotic manipulation. We argue that success with deformable objects requires robots to move beyond merely detecting failures to semantically classifying their physical cause and context (e.g., buckling, tearing) and responding accordingly. Our proof-of-concept experiments support this claim, showing that distinct failure modes produce unique tactile signatures (Fig.~\ref{fig:experiments}). However, we identify a key hardware bottleneck. These critical events are characterized by sub-newton forces ($\sim$0.1--0.5 N) that are often \textit{undetectable} by standard hardware, which is limited by coarse force resolution. We also show that first-principles models such as FEM are ill-suited for this task, as they are notoriously hard to calibrate and exhibit sim-to-real force errors (e.g., $\sim$1.5 N RMSE) an order of magnitude larger than the forces to be regulated. We conclude that SFA-DOM is a relevant and timely objective for robust, dexterous manipulation. Our analysis suggests that realizing it is feasible, but that it is not solely a software challenge. The main challenges and research directions we identify are (1) high-resolution, contact-proximal sensing, (2) high-bandwidth, high-resolution force actuation, (3) data-driven models trained on real, fine-contact data, and (4) control policies that explicitly exploit semantic failure labels ($s_t$) to enable robust, event-specific corrective actions.

\section{Acknowledgements}

This work was supported by the projects REMAIN - S1/1.1/E0111 (Interreg Sudoe Programme, ERDF),
 ANR-25-PERO-0003 PEPR Robotique - PC DRMI (Dexterous Robotic Manipulation for Industry), and by Engineering and Physical Sciences Research Council [grant number EP/W524633/1]. Part of this work was conducted at Instituto de Investigación en Ingeniería de Aragón, Universidad de Zaragoza, Spain.

\bibliographystyle{splncs04}
\bibliography{bibliography}

\end{document}